\documentclass[11pt,a4paper]{article}
\usepackage[hyperref]{ranlp2023}
\usepackage{times}
\usepackage{latexsym}

\usepackage{microtype}

\aclfinalcopy 

\usepackage[inline]{enumitem}

\usepackage{graphicx,booktabs}
\usepackage{subfig}
\usepackage{multirow}
\usepackage{float}
\usepackage[T1]{fontenc}
\usepackage{caption}

\title{Evaluating Generative Models for Graph-to-Text Generation}

\author{Shuzhou Yuan \and Michael Färber \\
        Karlsruhe Institute of Technology (KIT) \\ \texttt{\{shuzhou.yuan, michael.faerber\}@kit.edu}}

\date{}

\begin{document}
\maketitle
\begin{abstract}
Large language models (LLMs) have been widely employed for graph-to-text generation tasks. However, the process of finetuning LLMs requires significant training resources and annotation work. In this paper, we explore the capability of generative models to generate descriptive text from graph data in a zero-shot setting. Specifically, we evaluate GPT-3 and ChatGPT on two graph-to-text datasets and compare their performance with that of finetuned LLM models such as T5 and BART. 
Our results demonstrate that generative models are capable of generating fluent and coherent text, achieving BLEU scores of 10.57 and 11.08 for the AGENDA and WebNLG datasets, respectively. However, our error analysis reveals that generative models still struggle with understanding the semantic relations between entities, and they also tend to generate text with hallucinations or irrelevant information. As a part of error analysis, we utilize BERT to detect machine-generated text and achieve high macro-F1 scores.
We have made the text generated by generative models publicly available.\footnote{\url{https://github.com/ShuzhouYuan/Eval_G2T_GenModels}} 
\end{abstract}

\begin{figure*}[hbt!]

  \subfloat[\underline{Generate paper abstract from title, entities and graph:} \textbf{<title>} Significance-aware Hammerstein group models for nonlinear acoustic echo cancellation. \textbf{<entities>} non-linear preprocessor echo path hammerstein model \textbf{<graph>} \textbf{<H>} non-linear preprocessor \textbf{<R>} USED-FOR \textbf{<T>} echo path \textbf{<H>} preprocessor \textbf{<R>} EVALUATE-FOR \textbf{<T>} hammerstein model \textbf{<H>} hammerstein model \textbf{<R>} USED-FOR \textbf{<T>} echo path]{\includegraphics[width=0.5\textwidth]{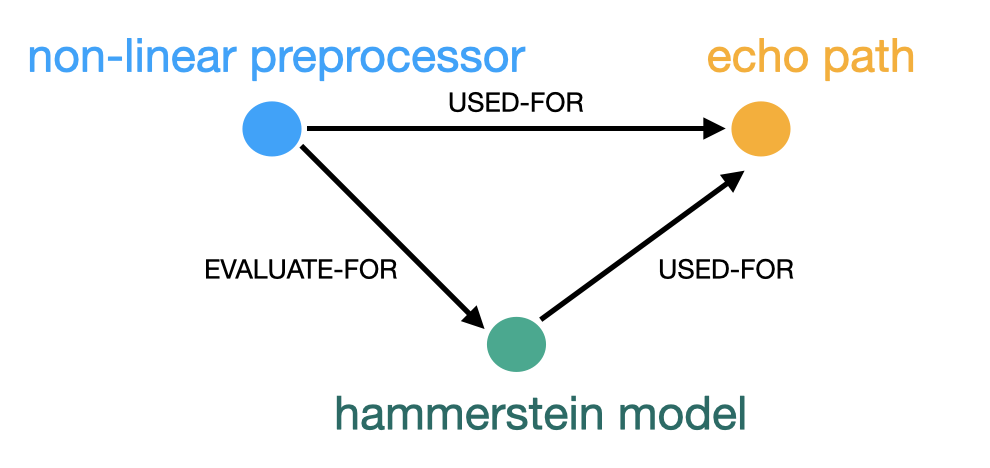}}
  \hfill
  \subfloat[\underline{Generate text from graph:} \textbf{<H>} Auburn Washington \textbf{<R>} is Part Of \textbf{<T>} Pierce County Washington \textbf{<H>} Pierce County Washington \textbf{<R>} country \textbf{<T>} United States ]{\includegraphics[width=0.48\textwidth]{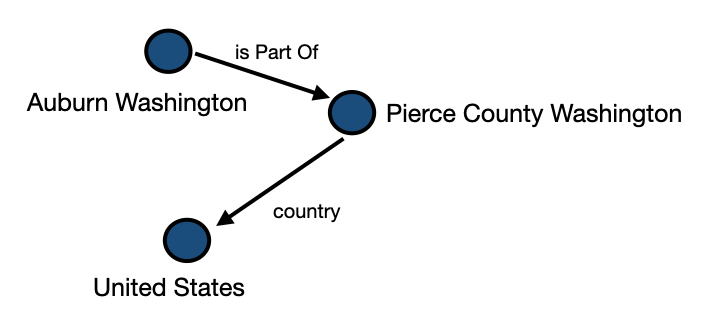}}
  \caption{Examples of graph structures, prompts and linearised graphs of (a) AGENDA and (b) WebNLG.}\label{examples}
\end{figure*}

\section{Introduction}

Graph-to-text generation is a subtask of data-to-text generation and natural language generation (NLG) \cite{gatt2018survey}. Its purpose is to generate fluent descriptive text based on the structure of a given graph (see Figure \ref{examples}). With the widespread use of graph structured data, this technique plays a crucial role in various natural language processing applications, including question answering, dialogue systems, and data augmentation \cite{he-etal-2017-generating, zhao-etal-2020-bridging, josifoski2023exploiting}. Previous research on model architectures has achieved significant performance on graph-to-text generation benchmarks \cite{koncel-kedziorski-etal-2019-text,ribeiro-etal-2020-modeling,zhao-etal-2020-bridging,li-etal-2021-shot-knowledge,ribeiro-etal-2021-structural}. In particular, \citet{ribeiro-etal-2021-investigating} achieved state-of-the-art performance by employing large pretrained language models and sufficient training data. However, the zero-shot setting for graph-to-text generation remains challenging due to the inconsistent input format (unstructured text vs. preformatted text) between pretraining and fine-tuning stages for large language models.

Recently, generative models such as GPT-3 \cite{brown2020language}, InstructGPT \cite{ouyang2022training}, and ChatGPT have gained tremendous attention in both the NLP research community and the general public. Researchers have evaluated these models on various NLP benchmarks in the zero-shot setting \cite{bang2023multitask,jiao2023chatgpt,ahuja2023mega}. However, their ability to process structured data, and in particular graph data, such as knowledge graphs, is understudied and worth being explored \cite{bang2023multitask}. Given the significant resources and annotations required for training graph-to-text generation models \cite{li-etal-2021-shot-knowledge}, utilizing a zero-shot setting could save training resources and prove advantageous for both economic and ecological reasons.

Previous approaches has come up with a neural pipeline to enable zero-shot for graph-to-text generation but didn't use generative models  \cite{kasner-dusek-2022-neural}. In contrast, our approach adopts the zero-shot setting by using prompts as instructions for generative models, specifically GPT-3 and ChatGPT \cite{brown2020language, ouyang2022training}. We evaluate the models' ability to translate graph data into fluent text using the test sets from two widely used graph-to-text generation datasets: WebNLG \cite{gardent-etal-2017-webnlg} and AGENDA \cite{koncel-kedziorski-etal-2019-text}. Following the method of \citet{ribeiro-etal-2021-investigating}, we represent the graph as a linearized sequence of text for input to the models (see Figure~\ref{examples}).

To assess the performance of the generative models, we conduct a comprehensive evaluation on each dataset. Employing machine translation metrics such as BLEU \cite{papineni-etal-2002-bleu}, METEOR \cite{banerjee-lavie-2005-meteor}, and ROUGE \cite{lin-2004-rouge} to the generated texts, we reveal that the generative models fall short of matching the quality achieved by state-of-the-art approaches. To identify patterns of mistakes made by the generative models, we perform error analysis by comparing the generated texts with the reference texts. Additionally, we fine-tune a BERT model to detect the machine-generated text. We make the texts generated by the models available on GitHub to facilitate future research on the analysis of machine-generated text and trustworthy AI.

In summary, our study aims to assess the performance of generative models in the zero-shot setting for graph-to-text generation using two distinct benchmarks. Our contribution lies in conducting a rigorous quantitative analysis of the results, shedding light on the effectiveness of generative models in this domain.

\section{Related Work}

\textbf{Graph-to-text generation.} Various efforts have been made to enhance graph-to-text generation using neural network models. They can be categorized into two main types: Graph Neural Network (GNN) based models and Language Model (LM) based models. GNN-based models typically employ a graph encoder to encode the graph structure \cite{beck-etal-2018-graph,marcheggiani-perez-beltrachini-2018-deep,damonte-cohen-2019-structural,koncel-kedziorski-etal-2019-text,ribeiro-etal-2019-enhancing,li-etal-2021-shot-knowledge}. In contrast, LM-based models do not rely on the graph structure but purely on the sequence of tokens in the text. As such, graphs have first been transformed into a linearized representation before being fed into LMs to generate coherent text \cite{harkous-etal-2020-text,ribeiro-etal-2021-investigating,ribeiro-etal-2021-structural}. Besides GNN and LM, previous works have also explored the use of Recurrent Neural Networks (RNNs) such as LSTM and GRU for graph-to-text generation \cite{song-etal-2018-graph,zhao-etal-2020-bridging,guo-etal-2020-cyclegt}. We follow the approach of \citet{konstas-etal-2017-neural} and other prior works by using a linearized graph as input for generative models.

\textbf{Generative Models.} Generative language models, such as GPT-3 \cite{brown2020language}, InstructGPT \cite{ouyang2022training}, and ChatGPT, have been designed to learn and generate natural language text. These models are based on the transformer decoder architecture \cite{NIPS2017_3f5ee243}, which enables them to handle large amounts of training data and perform zero-shot applications. While GPT-3 has made a significant breakthrough in text completion, InstructGPT and ChatGPT possess unique characteristics that align user intent with a conversational style. These models are trained using supervised fine-tuning and reward modeling, allowing them to generate high-quality responses that accurately reflect the user's needs and preferences. InstructGPT and ChatGPT are first fine-tuned on the GPT-3 model through supervised learning and then further trained using reinforcement learning based on human feedback. As demonstrated by \citet{ouyang2022training}, this approach substantially improves the model's performance on NLP benchmarks. Although there have been numerous reports and research evaluating the performance of generative models in various NLP applications such as summarization \cite{bang2023multitask}, machine translation \cite{jiao2023chatgpt}, and multilingual evaluation \cite{ahuja2023mega}, our work focuses on the generative models' capability to handle structured data.

\section{Dataset}

\begin{table}[tb]
		\caption{Statistics of test sets from AGENDA and WebNLG.}
		\label{statistics}
		\centering
		\begin{tabular}{lrr}
	\toprule		
    & \textbf{AGENDA} & \textbf{WebNLG}\\
    \midrule 
			Number of Instance & 1,000 & 1,862\\
			Average Input Tokens & 169 & 66 \\
   \bottomrule
		\end{tabular}
	\end{table}

 We evaluate generative models using the AGENDA and WebNLG datasets, as they are widely used in recent research on graph-to-text generation \cite{koncel-kedziorski-etal-2019-text,ribeiro-etal-2021-investigating,li-etal-2021-shot-knowledge} and as they represent different domains: scholarly domain and general domain (e.g., as given in Wikipedia). We focus on the test sets of AGENDA and WebNLG for our experiments, as the models do not require further training. In the following, we briefly describe the used datasets. 

\textbf{AGENDA.} Abstract GENeration DAtaset (AGENDA) is a dataset that pairs knowledge graphs with paper abstracts from scientific domains \cite{koncel-kedziorski-etal-2019-text}. The graphs in AGENDA were automatically extracted from the SciIE information extraction system \cite{luan-etal-2018-multi}. Each instance in AGENDA includes the title, entities, graph, and abstract of a paper. We use the title, entities, and graph as input for the models.

\textbf{WebNLG.} This dataset is a benchmark for mapping sets of RDF triples to text \cite{gardent-etal-2017-webnlg}. The RDF triples are subgraphs of the knowledge graph DBpedia \cite{10.1007/978-3-540-76298-0_52}, while the texts describe the graphs in one or a few sentences. The WebNLG challenge\footnote{\url{https://synalp.gitlabpages.inria.fr/webnlg-challenge/}} has released several versions of this dataset since 2017. In order to compare with previous work, we take the test data of WebNLG challenge 2017 for our experiments.

\section{Experiments}

\textbf{Data Preprocessing.} Since GPT-3 and ChatGPT require a sequence of text as input, we convert the graph structure into a linearized representation following \citet{ribeiro-etal-2021-investigating}. To assist the models in identifying the head, relation, and tail entities, we prepend \texttt{<H>}, \texttt{<R>}, and \texttt{<T>} tokens before the entities, as done in previous work \cite{harkous-etal-2020-text}. In the AGENDA dataset, each sample also includes a title and entities. Thus, we additionally add \texttt{<title>}, \texttt{<entities>}, and \texttt{<graph>} tokens (see Figure \ref{examples}). 

\textbf{Model Settings.} We use the GPT-3 model variant \texttt{text-davinci-003} and the ChatGPT model variant \texttt{gpt-3.5-turbo-0301} for our experiments. Each instance is treated as a single request, and the first response from the model is taken as the generated text. The prompt used for the models plays a significant role as it serves as the task description and directly influences the content of the generated text. Previous work designed prompts by asking ChatGPT \cite{jiao2023chatgpt}. Following their approach, we ask ChatGPT to provide prompts: ``Please provide prompts or templates for graph-to-text generation:''. Since AGENDA and WebNLG have different data structures, we use the prompt ``Generate paper abstract from title, entities, and graph:'' for AGENDA. For WebNLG, we use the prompt ``Generate text from graph:''. 
We expect that in this way the generated text fits the format of a scientific paper abstract better for AGENDA, while the models generate texts in open domain for WebNLG. 

	\begin{table*}[tb]
 		\caption{Results on AGENDA. 
			\label{results_agenda}}
		\centering
		\begin{tabular}{lrrrrr}
  \toprule
			\textbf{Model} & \textbf{BLEU}$\uparrow$ & \textbf{METEOR}$\uparrow$ & \textbf{RougeL}$\uparrow$ & \textbf{Chrf++}$\uparrow$& \textbf{BLEURT}$\uparrow$\\
   \midrule
            T5\textsubscript{large} \cite{ribeiro-etal-2021-investigating}  & 22.15 & 23.73 & - & - & -13.96 \\
            BART\textsubscript{large} \cite{ribeiro-etal-2021-investigating}  & \textbf{23.65} & \textbf{25.19} & - & - & \textbf{-10.93} \\
             GPT-3  & 8.34 & 14.88 & 24.99  & 41.42 & -32.54 \\
			ChatGPT  & 10.57 & 17.02 & \textbf{25.22}& \textbf{45.86} & -28.05\\
   \bottomrule
		\end{tabular}
	\end{table*}

	\begin{table*}[tb]
 		\caption{Results on WebNLG.
			\label{results_webnlg}}
		\centering
		\begin{tabular}{lrrrrr}
			\toprule 
   \textbf{Dataset} & \textbf{BLEU}$\uparrow$ & \textbf{METEOR}$\uparrow$ & \textbf{RougeL}$\uparrow$ & \textbf{Chrf++}$\uparrow$& \textbf{BLEURT}$\uparrow$\\ \midrule
            T5\textsubscript{large}\cite{ribeiro-etal-2021-investigating}  & \textbf{59.70} & \textbf{44.18} & - & \textbf{75.40} & - \\
            BART\textsubscript{large}\cite{ribeiro-etal-2021-investigating}  & 54.72 & 42.23 & - & 72.29 & -\\
			 GPT-3 & 20.36 & 26.95 & \textbf{45.64} & 57.95 & \textbf{13.39} \\
			 ChatGPT & 11.08 & 23.89 & 35.87 & 48.75 & -10.99 \\
    \bottomrule
		\end{tabular}
	\end{table*}

 \textbf{Baseline.} Similar to our experimental methodology, \citet{ribeiro-etal-2021-investigating} finetuned T5 and BART using linearized graphs as input and generated descriptive texts. Therefore, we consider their findings as the baseline for comparison with our own experiments.
 
\textbf{Evaluation.} Following related work, we implement a thorough evaluation with metrics BLEU \cite{papineni-etal-2002-bleu}, METEOR \cite{banerjee-lavie-2005-meteor}, RougeL \cite{lin-2004-rouge} and Chrf++ \cite{popovic-2017-chrf}. Additionally, to assess the semantic meaning and coherence of the generated text, we employ BLEURT \cite{bleurt}, a metric that evaluates not only the surface match of n-grams but also the semantic representation extracted from a pretrained BERT \cite{devlin-etal-2019-bert} model. 

\subsection{Results}

Our results are summarized in Table \ref{results_agenda} and \ref{results_webnlg}. As comparison, we take the results from \citet{ribeiro-etal-2021-investigating}, which are achieved by finetuned BART and T5.

The results obtained from AGENDA demonstrate that finetuned BART and T5 models outperform generative models in terms of state-of-the-art performance. Both T5 and BART achieve BLEU scores exceeding 20, while GPT-3 only attains a BLEU score of 8.34 and ChatGPT achieves 10.57. Consistently, other evaluation metrics align with the BLEU scores, further highlighting the limited performance of generative models without finetuning. Notably, ChatGPT exhibits a slightly improved performance compared to GPT-3 on the AGENDA benchmark. Analysis of the results reveals that ChatGPT consistently outperforms GPT-3 across all metrics, showcasing a 2.23 higher BLEU score, a 2.14 higher METEOR score, a 0.23 higher RougeL score, a 4.44 higher Chrf++ score, and a 4.49 higher BLEURT score.

Examining the results from WebNLG, it becomes evident that fine-tuned T5 and BART models consistently outperform generative models without fine-tuning. Notably, both T5 and BART achieve BLEU scores exceeding 50, whereas generative models only attain a BLEU score of 11.08 for ChatGPT and 20.36 for GPT-3. Surprisingly, GPT-3 outperforms ChatGPT on the WebNLG benchmark with a BLEU score that is 9.28 higher, a METEOR score that is 3.06 higher, a RougeL score that is 9.77 higher, and a Chrf++ score that is 9.20 higher. The primary reason for this difference is that ChatGPT tends to produce hallucinations easily and generates longer text. We provide further elaboration on two examples in Section \ref{e_a}.

\section{Error Analysis\label{e_a}}

 \begin{table*}[h!]
               \caption{Paper abstract generated by GPT-3 and ChatGPT, incorrect texts are marked with \textcolor{red}{red}.
               \label{agenda_error}}
    \centering
		\begin{tabular}{lp{12cm}}
			\toprule  \textbf{AGENDA} & \textbf{Example}\\
			\midrule 
   Input & Generate paper abstract from title, entities and graph: <title> Estimation of semantic case of Japanese dialogue by use of distance derived from statistics of dependency . <entities> statistics of dependent noun-particle-verb triples atr dialogue corpus measure of distance semantic cases noun-particle-verb triples consistency rates single-case clusters clustering analysis correlation analysis estimation accuracies accuracy <graph>  <H> statistics of dependent noun-particle-verb triples <R> USED-FOR <T> measure of distance <H> noun-particle-verb triples <R> PART-OF <T> atr dialogue corpus <H> semantic cases <R> PART-OF <T> atr dialogue corpus <H> accuracy <R> EVALUATE-FOR <T> correlation analysis\\
   \midrule 
			  Reference & In an attempt to estimate the semantic cases for noun-particle-verb triples in the atr dialogue corpus, the authors propose a measure of distance based on statistics of dependent noun-particle-verb triples. A clustering analysis of all the triples in the corpus was conducted using the measure of distance. Competence of the proposed measure of distance is verified by examination of the distribution of the single-case clusters. By use of the score derived from the measure of distance of the training corpus, the authors conducted the estimation of the correct semantic case for a given noun-particle-verb triples in the test corpus. The result remarkably differentiates the particles with respect to the estimation accuracies. For instance, particle `wo' has accuracies over 80 \%, while `de' has accuracies less than 40\%. The correlation analysis between the accuracy and the consistency rates indicates that the particles of higher consistency have also tendencies to higher accuracies. \\\midrule 
            GPT-3 & This paper aims to estimate semantic cases of Japanese dialogue using a measure of distance derived from statistics of dependent noun-particle-verb triples from the ATR dialogue corpus. Clustering analysis and \textcolor{red}{correlation analysis} are employed to examine the consistency rates of single-case clusters, which is then used \textcolor{red}{to evaluate} the estimation \textcolor{red}{accuracies} of the distance measure.\\\midrule
             ChatGPT & This paper proposes a method to estimate the semantic case of Japanese dialogue using a measure of distance derived from the statistics of dependent noun-particle-verb triples. The authors use the ATR dialogue corpus and apply clustering and \textcolor{red}{correlation analysis to evaluate the accuracy} of their method. They report consistent rates for single-case clusters and show that their method achieves high estimation accuracies. The study demonstrates the potential of using statistics of dependency for analyzing Japanese dialogue.\\
             \bottomrule
             	\end{tabular}
	 \end{table*}

\begin{table*}[tb!]
          \caption{Descriptive text generated by GPT-3 and ChatGPT, incorrect texts are marked with \textcolor{red}{red}, hallucinations are marked with \textcolor{blue}{blue}.}
			\label{webnlg_error}
	\centering
		\begin{tabular}{cp{12cm}}
			\toprule \textbf{WebNLG} & \textbf{Example}\\
			\midrule 
   Input & Generate text from graph: <H> Abner W. Sibal <R> battles <T> World War II <H> World War II <R> commander <T> Joseph Stalin\\ 
   \midrule
			  Reference & Joseph Stalin was a commander in World War II in which Abner W Sibal also fought.\\
     \midrule
            GPT-3 & Joseph Stalin was the commander of World War II and \textcolor{red}{Abner W. Sibal was one of the battles} fought during the war.\\ 
            \midrule 
             ChatGPT & Abner W. Sibal was a prominent figure who faced great challenges during his time as a soldier in World War II. \textcolor{blue}{He fought bravely as a commander, leading his troops with dedication and determination. His efforts were critical to the success of the war effort and his name became synonymous with bravery and patriotism.  During World War II, the world was gripped in the fight against tyranny and oppression. Abner W. Sibal was one of the many who answered the call to serve his country. Through his leadership skills and strategic thinking, he was able to navigate the treacherous waters of war and emerge victorious.  However, the fight against the Axis powers was not just a battle against military might. It was a battle against the ideologies that fueled the enemy's aggression. As a commander, Sibal had to navigate these complexities and ensure that his troops remained committed to the cause of freedom.  Throughout the war, Sibal faced many challenges and obstacles. But he remained steadfast and only grew more determined to see the war through to its ultimate conclusion. His efforts were recognized and honored, not just by his own country but also by leaders around the world. Even the likes of Joseph Stalin, renowned commander and leader of the Soviet Union, recognized the critical role Sibal played in the victory against fascism.  Today, Abner W. Sibal's name remains a testament to the bravery and dedication of those who fought in World War II. His legacy serves as an inspiration to future generations, reminding them of the importance of standing up against tyranny and oppression wherever it may arise.}\\
             \bottomrule
             \end{tabular}
	\end{table*}

We observe that the texts generated by generative models contain errors following similar patterns. In Table \ref{agenda_error} and Table \ref{webnlg_error}, we show two examples from AGENDA and WebNLG.

As shown in the example of Table \ref{agenda_error}, generative models are capable of generating  fluent paper abstracts based on the provided input, but they contain errors that contradict the facts represented by the graph structure. For instance, the triple ``accuracy <R> EVALUATE-FOR <T> correlation analysis'' in the example graph is wrongly generated by GPT-3 as ``which (correlation analysis) is then used to evaluate the estimation accuracies''. Also, ChatGPT describes it as ``(The authors) apply clustering and correlation analysis to evaluate the accuracy''. In both cases, the models reverse the relationship between ``accuracy'' and ``correlation analysis''. 

Furthermore, ChatGPT ignores the relations ``PART-OF'' and ``USED-FOR'' in the input. This suggests that generative models still face challenges in comprehending the semantic relations between entities, which can result in the generation of inaccurate text. 

 	\begin{table*}[tbh!]
  		\caption{Statistics of ChatGPT generated datasets}
			\label{our_data}
		\centering
		\begin{tabular}{lcc}
  \toprule 
			\textbf{Dataset} & \textbf{Machine-generated instance} & \textbf{Human-written instance}\\
   \midrule 
			AGENDA & 1000 & 1000\\
			WebNLG & 1862 & 4894 \\
             All  & 2862 & 5894 \\
             \bottomrule
		\end{tabular}
	\end{table*}

	\begin{table}[tb!]
 		\caption{Results of BERT to detect GPT-3 and ChatGPT generated text.}
			\label{bert}
		\centering
  \resizebox{\linewidth}{!}{%
		\begin{tabular}{lrr}
  \toprule 
		 Model & \textbf{Accuracy} & \textbf{Macro F1}\\
   \midrule 
		 GPT-3\textsubscript{AGENDA}& 98.00 & 98.00\\
            ChatGPT\textsubscript{AGENDA} & 100 & 100\\ \midrule 
			GPT-3\textsubscript{WebNLG} & 91.64 & 89.25 \\
            ChatGPT\textsubscript{WebNLG} & 96.82 & 95.75\\ \midrule 
             GPT-3\textsubscript{All}  & 93.55 & 92.38 \\
             ChatGPT\textsubscript{All}  & 96.40 & 95.82 \\
             \bottomrule
		\end{tabular}
  }
	\end{table}

While generating paper abstracts is complex and challenging, generating short descriptive text from a knowledge graph is relatively more straightforward. As LLMs, GPT-3 and ChatGPT are trained on large corpora and thus already contain world knowledge. In the example provided in Table \ref{webnlg_error}, generative models generate descriptive text fluently based on the structured input. However, GPT-3 produces text with incorrect facts. For instance, ``Abner W. Sibal'' is the name of a person, but it is recognized as the name of a battle by GPT-3. While GPT-3 produces text with incorrect facts, it is worth noting that ChatGPT-generated text not only covers the input information, but also contains redundant messages from its internal knowledge (hallucinations). Furthermore, the sentence ``Abner W. Sibal was a prominent figure who faced great challenges'' generated by ChatGPT has made a subjective judgement about the character and may cause unnecessary bias to potential users.

To investigate the difference between model-generated texts and human-written reference texts, we create datasets containing both types of text and finetune a pretrained BERT model for a binary text classification task. The statistics of our datasets are presented in Table \ref{our_data}.

We create several datasets for AGENDA, WebNLG, and a combined dataset containing both AGENDA and WebNLG examples. The training and test sets are split in an 80:20 ratio. We fine-tune BERT for five epochs using the AdamW optimizer \cite{loshchilov2019decoupled}. As shown in Table \ref{bert}, BERT achieves high scores across all datasets. This demonstrates that generative models generate text that follows similar patterns, and a state-of-the-art text classifier can easily distinguish between them.

\section{Conclusion}

In this paper, we explored the capabilities of generative models in generating coherent text from structured data, focusing on two benchmarks: AGENDA and WebNLG. To achieve this, we adopted the linearized graph representation approach employed in prior work. Leveraging the zero-shot ability of language models, we prepended the prompt to the input text as an instruction for both GPT-3 and ChatGPT. We conducted a comprehensive evaluation using various metrics. Our findings reveal that generative models fall short of surpassing previous models that have been trained and finetuned on large volumes of training data. These results highlight the limitations of generative models in achieving state-of-the-art performance in graph-to-text generation tasks.

Furthermore, we conducted an error analysis of the text generated by the models. The generative models struggle in capturing the relationships between entities and often produce unrelated information, leading to hallucinations. To further investigate the machine generated text, we employ finetuned BERT to conduct a text classification task. BERT achieves high F1 scores in distinguishing between machine-generated text and human-written text. Our study provides extensive evaluation of generative models for graph-to-text generation. Future work should focus on refining machine-generated text and reducing hallucinations for graph-to-text generation by using generative models.

\section{Ethical Consideration and Limitation}

We observe that generative models may generate text containing fake facts or offensive content. And the datasets we collected may also contain incorrect or offensive statements. We do not support the views expressed in the machine generated text, we merely venture to analyze the machine generated text and provide an useful resource for future research.

As the limitation of this work, we found out that the reproducibility of GPT-3 and ChatGPT is questionable. The models often return different response from same request, which makes our results hard to reproduce and the brings randomness to the evaluation scores.

\section*{Acknowledgements}
We thank the anonymous reviewers for their helpful comments. We also would like to thank Nicholas Popovic for his feedback on this work.

\bibliographystyle{acl_natbib}
\bibliography{ranlp2023}
\clearpage


\end{document}